\documentclass[letterpaper, 10 pt, conference]{ieeeconf}  % Comment this line out if you need a4paper

\IEEEoverridecommandlockouts                              % This command is only needed if 
\title{
\LARGE
\bf
Scale Invariant Semantic Segmentation with RGB-D Fusion
}

\usepackage[backend=bibtex,style=ieee,natbib=true]{biblatex}
\bibliography{example}
\usepackage{comment}
\usepackage{graphicx}
\usepackage{float}
\usepackage{flushend}
\newcommand{\etal}{\textit{~et al.~}}

\author{Mohammad Dawud Ansari $^{1}$, Alwi Husada$^{2}$ and Didier Stricker $^{1}$% <-this % stops a space
\thanks{$^{1}$German Research Center for Artificial Intelligence (DFKI GmbH), Germany. {\tt\small\{dawudmaxx@gmail.com, didier.stricker@dfki.de\}}}%
\thanks{$^{2}$University of Kaiserslautern, Germany. {\tt\small husada@rhrk.uni-kl.de}}%
}

\begin{document}

\maketitle
\thispagestyle{empty}
\pagestyle{empty}

\begin{abstract}

In this paper, we propose a neural network architecture for scale-invariant semantic segmentation using RGB-D images. 
We utilize depth information as an additional modality apart from color images only. 
Especially in an outdoor scene which consists of different scale objects due to the distance of the objects from the camera.
The near distance objects consist of significantly more pixels than the far ones.
We propose to incorporate depth information to the RGB data for pixel-wise semantic segmentation to address the different scale objects in an outdoor scene.
We adapt to a well-known DeepLab-v2(ResNet-101) model as our RGB baseline.
%
%% We add a branch to the baseline model to extract depth information and fusion block to fuse features from both branches, before passing the fused feature maps to the  Pyramid Pooling layers and later for final prediction.
Depth images are passed separately as an additional input with a distinct branch. The intermediate feature maps of both color and depth image branch are fused using a novel fusion block. 
Our model is compact and can be easily applied to the other RGB model. 
We perform extensive qualitative and quantitative evaluation on a challenging dataset Cityscapes.
The results obtained are comparable to the state-of-the-art. 
Additionally, we evaluated our model on a self-recorded real dataset.
For the shake of extended evaluation of a driving scene with ground truth we generated a synthetic dataset using popular vehicle simulation project CARLA.
The results obtained from the real and synthetic dataset shows the effectiveness of our approach.

\end{abstract}

\section{INTRODUCTION}
\begin{figure*}[!htpb]
\begin{center}
\includegraphics[width=1\textwidth]{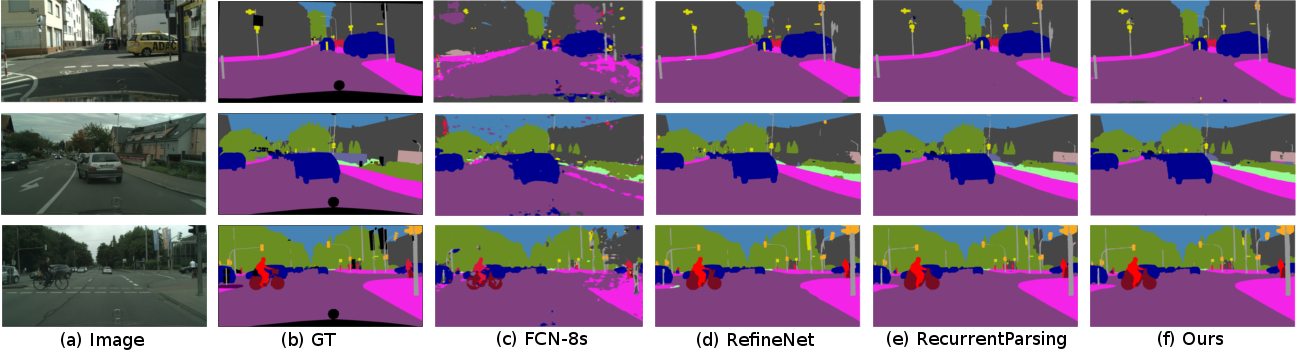}
\end{center}
\caption{Qualitative results of our model compared to FCN-8s \cite{fcn}, RefineNet \cite{refineNet}, RecurrentParsing \cite{rsp} and DeepLab-v2(ResNet-101)\cite{deeplabv2} on Cityscapes dataset (validation set).}
\label{fig:city_all}
\end{figure*}
Deep Convolutional Neural Networks (DCNNs) have shown remarkable accuracy for computer vision tasks like object classification \cite{krizhevsky2012imagenet,sermanet2013overfeat,simonyan2014very,szegedy2015going} and object detection \cite{girshick2014rich,erhan2014scalable,ren2017faster,he2016deep,liu2016ssd}.
%
%Many modern state-of-the-art approaches can even out-perform humans.
%
In the era of autonomous driving, classification and detection are not sufficient informations that can guide a vehicle in an unknown environment autonomously.
As an example, such a system must be able to detect pedestrian from car and house and so on.
In Computer Vision community applies for this task dense per-pixel classification called as pixel-wise semantic segmentation. 
%
%The goal of semantic segmentation task is to label each pixel in an image to have semantic meaning.
%
The goal of semantic segmentation is to recognize and understand which pixels level belong to which object in the image.
Semantic segmentation gives benefits to various applications such as robotic vision,  autonomous cars,  medical imaging data \cite{kayalibay2017cnn} etc.
Earlier approaches \cite{refineNet,pspnet, deeplabv2, deeplabv3, rsp} provide reasonable accuracy. 
However, this task is difficult with implicit problems such as illuminations, occlusions, cluttered background and multi-scale of objects.
Earlier multi-scale problem is solved by employing a network which takes input images with multiple resolutions and later aggregates the feature maps \cite{papandreou2015modeling,chen2016attention,kokkinos2015pushing}.
Since the computation is performed parallely for different scale, the overall computation cost is higher, in terms of memory requirement and computation power.
Other alternatives to solve multi-scale problem are proposed by \cite{deeplabv2,deeplabv3}.
Depth information can help to solve the problem of ambiguity in the scale \cite{rsp}.
Additionally, contextual information can be obtained by parallel pyramid pooling network as proposed in \cite{pspnet}.
We merge multi-scale feature generation along with contextual information and the depth information to tackle the problem of scale in driving scenes.
We summarize the contributions of the paper as follows:
\begin{itemize}
\item %1 - depth branch
We propose a new depth branch which takes the input as a raw depth image. 
The input layer of the neural network is modified to accept single channel input. 
Additionally, the resolution of diminishing activation map is retained similar to RGB branch utilizing dilated convolution. 
The intermediate full resolution feature maps from depth branch are merged with the RGB branch using a novel fusion block. 
Depth information resolves the ambiguity in scale change of similar objects.
\item 
We describe a scale-invariant pyramid pooling neural network model for semantic segmentation, which can cope with difficult scale changes for similar objects. 
Smaller dilation rate is used in every level of the pyramid pooling network which majorly focuses on the smaller objects.
However, to maintain the invariance for bigger objects, we adapt Global Average Pooling, which is also essential to keep the contextual information.
\item 
We describe a simple generation and utilization of a synthetic dataset with ground truth using vehicle simulation project CARLA \cite{carla}. 
An additional dataset with ground truth can be used to increase invariance in the neural network, and for extensive non-biased evaluation.
\item 
An extensive quantitative evaluation is performed on Cityscapes \cite{cityscapes} and synthetic driving dataset CARLA \cite{carla}. 
Additionally, we perform the qualitative evaluation on a self-generated real Zed dataset. 
The results obtained using our proposed model are promising for difficult scale changes of the similar object, which is also comparable to the state-of-the-art.
\end{itemize}

\section{RELATED WORKS}
%\subsection{RGB Models}
%
%
Over the past few years, the breakthroughs of Deep Learning in images classification were quickly transferred into the semantic segmentation task.
The introduction of Fully Convolutional Network \cite{fcn} which modifies the last fully connected layer to spatial output.
This enables for solving pixel-wise semantic segmentation that required the output label map to have same spatial size as the input.
The issues in the deep fully convolutional network is a set of stride in convolution and maximum/average pooling.
These operations downsample the spatial size of feature maps, causing the diminishing response.
Several approaches have been proposed to remedy the issue.
Shelhamer \etal \cite{fcn} upsample the feature maps from top layers and concatenate it with the feature maps from intermediate layers.
Eigen and Fergus \cite{eigenfergus} cascade multi-scale Deep Convolutional Neural Network (DCNN).
They concatenate fine-details result with the coarse input and use it as an input to the next DCNN.
Noh \etal \cite{noh}, use encoder-decoder architecture.
They apply deconvolution also known as transposed convolution in the decoder part.
Another approach is proposed by Chen \etal \cite{deeplabv2}, which we use as a baseline.
This method is referred as DeepLab-v2(ResNet-101).
They introduce "atrous convolution" to adjust receptive fields size by inserting "hole" in the filter.
It expands the receptive field without adding extra parameters.
A paper from Wang \etal \cite{DUC} proposed different method to upsample spatial size of the top feature maps.
Instead of using deconvolution \cite{noh, fcn} or bilinear interpolation \cite{deeplabv3, rsp}, they introduce Dense Upsampling Convolution (DUC), in which they arrange the top layer shape to have multiple channels of label map.
%

%\subsection{RGB-D Models}
%
Several Deep Convolution Neural Network (DCNN) models are proposed which combines depth information with color information for semantic segmentation \cite{fusenet, RDFNet, locality-sensitive}.
The fusion strategy affects the overall accuracy of the model.
We experiment our model with different fusion strategy and use the optimal one.
Hazirbas \etal \cite{fusenet} proposed an encoder-decoder type model that use two branches in encoder part to extract feature maps of RGB and depth images separately. They sum both feature maps which is later fed to the decoder part to get the final prediction.
We adapt similar methodology in our approach with additional modifications.
Cheng \etal \cite{locality-sensitive} introduces a gated fusion method.
It consists of concatenation, convolution and sigmoid layer.
They concatenated the top feature maps from RGB and depth branch, and then the resulting features are convolved with $3 \times 3$ filters followed by sigmoid layer to regularize the features. The outputs of sigmoid are used to weigh the contribution of depth and RGB features.
Park \etal \cite{RDFNet} extends RefineNet \cite{refineNet} architecture to integrate depth information.
They propose Multi-modal Feature Fusion (MMF), which composed of the same component as RefineNet block but it accepts RGB and depth feature maps as inputs.
Kong and Fowlkes \cite{rsp} proposed  another direction of incorporating depth information to the RGB-based model.
Instead of using different branches for feature extraction, their approach estimates depth information, and it then used for gating the size of pooling field in convolution model.
They trained their gating method with Recurrent Neural Network (RNN) strategy.
Our work is motivated by the gating mechanism as it resembles size invariance in the feature maps.
This is useful for generating intermediate feature maps which are scale invariant.
\begin{figure*}[!htpb]
\begin{center}
	\includegraphics[width=1\textwidth]{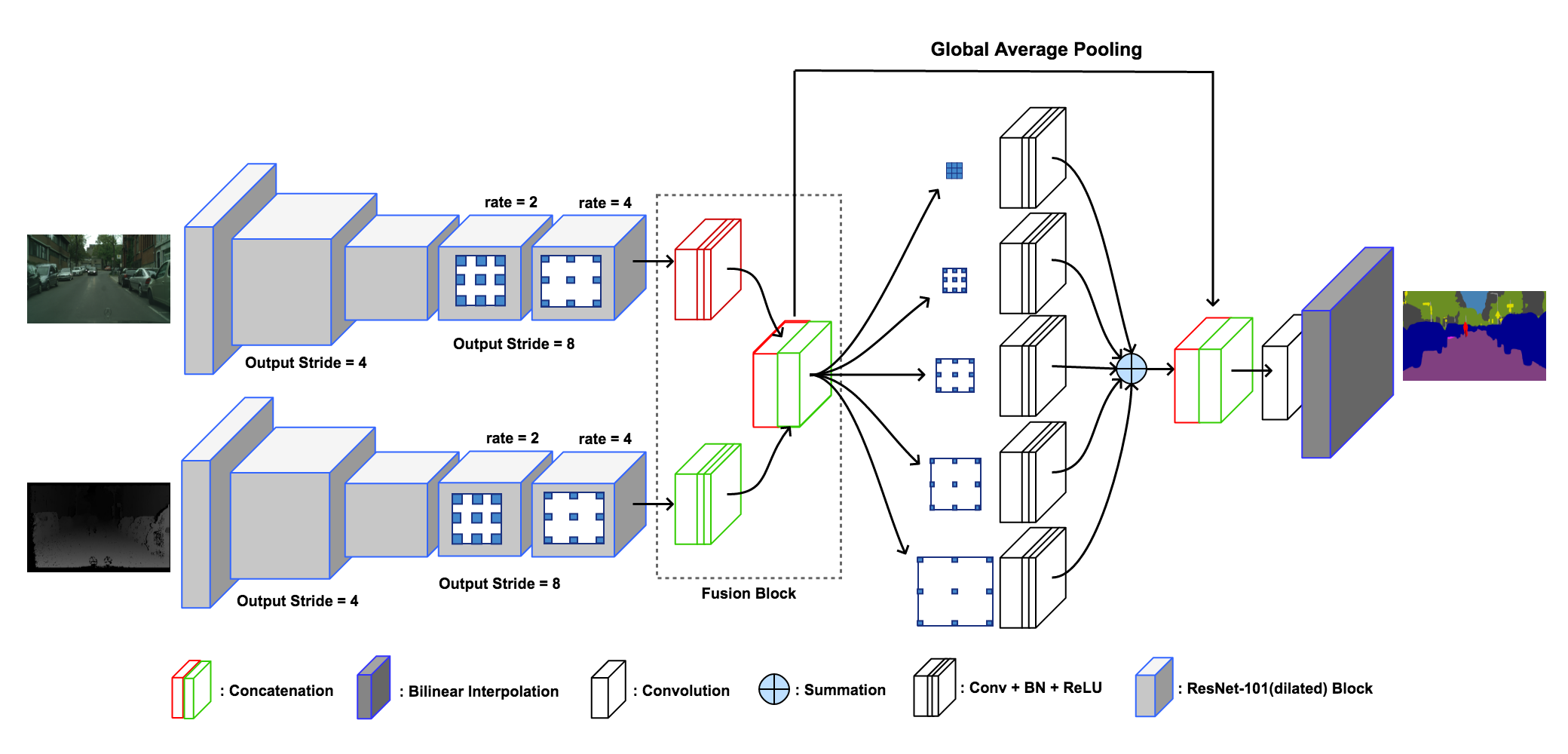}
\end{center}
  \caption{Overview of our proposed architecture. The model mainly composed of two branches of features generation, Fusion Block Layer and Pyramid Pooling. 
  We use Atrous ResNet-101 to extract robust features from both RGB and depth images.
  We then combine feature maps from both branches in the Fusion Block, which is passed to the Pyramid Pooling and later to final prediction. Additionally, Global Average Pooling is applied to get contextual information.}
\label{fig:models}
\end{figure*}
%
%%A recently published paper by Ansari \textit{et al.}\cite{scalenet} estimates quantized depth to adjust receptive field size through adaptive pooling.
%
%

\section{PROPOSED ARCHITECTURE}
The proposed model architecture is motivated from DeepLab-v2(ResNet-101) \cite{deeplabv2}. 
This model has two major parts. The first part is referred as ResNet-101 \cite{resnet} for feature maps generation and the second part is referred as Pyramid Pooling.
This combined network is named Atrous Spatial Pyramid Pooling (ASPP).
Our model can be seen in Figure \ref{fig:models}.
In contrast to DeepLab-v2(ResNet-101) \cite{deeplabv2} our model has a parallel branch that accepts depth image as input.
The features are generated from color as well as depth image separately. 
We then combine feature maps from both branches in the fusion block, which is passed to the ASPP and later to final prediction.
The ASPP consists of five parallel convolutions + Batch Normalization + ReLU blocks with different dilation rate.

\subsection{RGB-D Architecture}
We will briefly review DeepLab-v2(ResNet-101) architecture proposed by \cite{deeplabv2}.
One implicit problem in Deep Convolution Neural Network (DCNN) for Semantic Segmentation is consecutive maximum/average pooling and striding in 
convolutions operation that reduce spatial input dimension into significantly smaller feature maps.
This downsample factor is about $32$ with the actual input dimension. 
To overcome this issue Chen \etal \cite{deeplabv2} proposed ’atrous’ convolution also known as dilated convolution. 
The output signal $o\{i\}$ obtained after applying a dilated convolution in a 1D signal is given as:
\begin{equation}
o[i] = \sum_{l=1}^{L} x[i + r \cdot l]f[l],
\label{eq::atrous}
\end{equation}
where $x[i]$ is an input signal, $f[l]$ is an filter that has length $L$ and $r$ is a dilation rate. 
Noting that, when $r = 1$ is a standard convolution. 
In a 2D convolutional operation, atrous convolution can be seen as inserting 'hole' to the convolution filter (inserting zeros between two neighboring pixels in the filter). 
By defining the dilation rate $r$, it allows us to modify the size of receptive fields without changing the filter size. 
In addition, it also lets us control the spatial output size of feature maps of convolutions operation. 
In DeepLab-v3 \cite{deeplabv3} the spatial output size of feature maps is denoted as \textit{output\_stride}, which can also be termed as the reduction factor of the spatial input size to produce the desired output. 
For example, in original ResNet-101 model, the \textit{output\_stride} is $32$ \cite{resnet}. 
In order to obtain a bigger spatial dimension of the output feature maps, assuming the \textit{output\_stride} to be $16$, the stride of the last pooling and convolution layer is set to $1$ and modify the dilation rate of the subsequent convolution layers to $2$.
The feature maps from the last ResNet-101 layer are fed to the Atrous Spatial Pyramid Pooling (ASPP). 
It consists of several parallel filters with different dilation rate to exploit different scale of features. 
In DeepLab-v2(ResNet-101) \cite{deeplabv2}, the ASPP composed of four different filters with dilation rate \{6, 12, 18, 24\}. They are summed together before upsampled by \textit{output\_stride} factor with bilinear interpolation. 
In our model, we use  $\textit{output\_stride} = 8$. 
The repetitive convolution operation causes the diminishing of size of the features maps. 
An alternative to this operation is the use of $\textit{output\_stride} = 1$ along with dilation rate and modifying the stride of all the convolution layer to obtain same size feature maps. 
However, this operation is costly in terms of memory and training time. 
Thus $\textit{output\_stride} = 8$ is a reasonable choice to deal with the trade-off between memory usage and accuracy. 
This is the reason for setting the $\textit{output\_stride} = 8$ in our model. 
To handle depth data, we add an additional dilated ResNet-101 \cite{resnet} branch with $\textit{output\_stride} = 8$. 
We modify the first convolution layer to accept one channel image instead of three channels (RGB).
The rest of the layers in depth branch are same as its RGB counterpart. 
The explanation of this part of the model is shown in Figure \ref{fig:models}.

\subsection{Fusion Block} 
One trivial way to integrate depth information into existing RGB-based model, is by stacking both images as a single 4-channel RGB-D image and modifying the first convolution layer to accept a 4-channel input \cite{couprie}. 
Furthermore, Hazirbas \etal \cite{fusenet} showed that stacking RGB-D input produces less discriminant features than when fusing RGB and depth feature maps. 
The intuitive idea of fusing RGB and depth feature maps can be seen in the term of neuron-wise activation. 
For a given pixel, the activation maps from color and depth branch compliment each other simultaneously thereby producing more accurate segmentation. 
Motivated from this idea along with Hazirbas \etal \cite{fusenet} we propose fusion block which is composed by convolution followed by
%
%\clearpage
%
\begin{figure*}[h]
\begin{center}
	\includegraphics[width=1\textwidth]{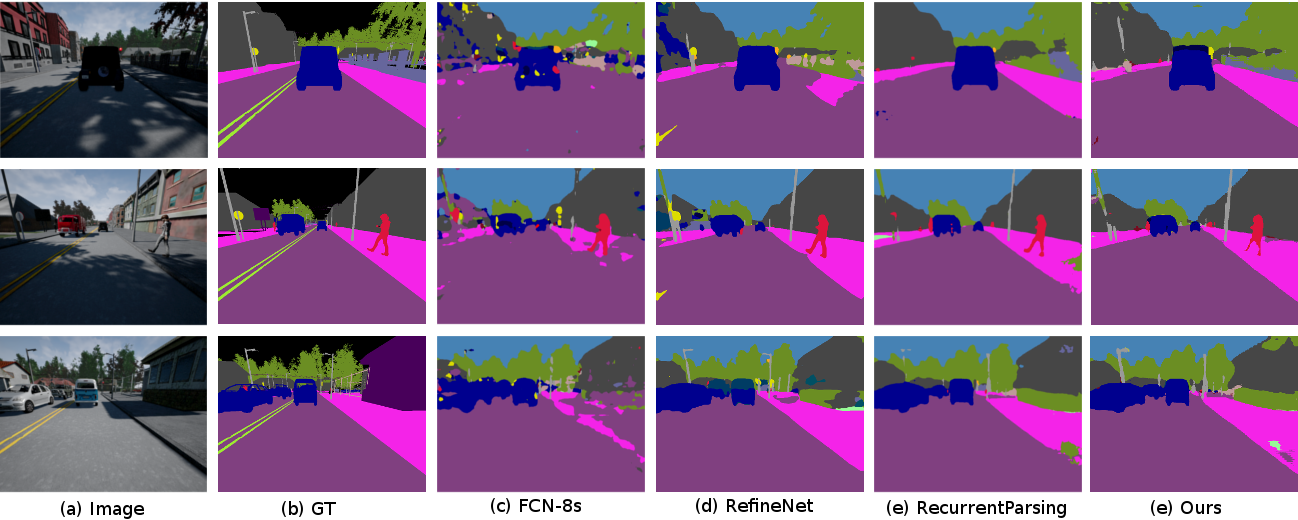}
\end{center}
  \caption{Qualitative results of our model compared to FCN-8s \cite{fcn}, RefineNet \cite{refineNet}, RecurrentParsing \cite{rsp} and DeepLab-v2(ResNet-101)\cite{deeplabv2} on CARLA dataset.}
\label{fig:carla_all}
\end{figure*}
%
%!htpb
\begin{figure*}[h]
\begin{center}
	\includegraphics[width=1\textwidth]{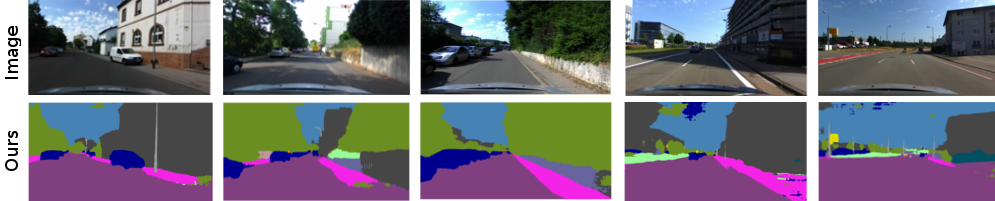}
\end{center}
  \caption{Qualitative results of our model on self-recorded Zed dataset.}
\label{fig:val_zed}
\end{figure*}
summation or concatenation. 
Given feature maps of RGB and depth from the last Atrous ResNet-101 with dimension $H  \times W \times 2048$ where H and W are height and width of the feature maps, we reduce the dimension to be $H \times W \times 512$. 
We then fuse both feature maps to get more discriminant features representation, which are then fed to the pyramid pooling network. 
We experiment with sum and concatenation for fusing the feature maps. 
The results show that concatenation produce better accuracy.
The explanation of this part of the model is shown in Figure \ref{fig:models}.

\subsection{Pyramid Pooling}
Many earlier approaches were proposed which provides significant accuracy with single object without much variation in the object's scale \cite{fcn, noh, fusenet}.
This creates an implicit issue for segmenting multi-scale objects in a scene.
Especially small objects which are far away from the camera. 
%
%% These far distant objects will create ambiguity with their closed distant similar objects due to scale-invariant features generated in DCNN.
%
Motivated from  \cite{fusenet, deeplabv2}, we intend to solve this issue by incorporating depth information and applying pyramid pooling.
In contrast, we apply five parallel convolutions + Batch Normalization + ReLU blocks with different dilation rate.
We modified $2^{nd}$ till $5^{th}$ pyramid level convolution operations with $3 \times 3$ filters with dilation rate \{2, 4, 8, 16\} respectively. 
The $1^{st}$ pyramid level convolution operation uses $1 \times 1$ filters without dilation rate as in \cite{deeplabv3}.
With small dilation rate the contextual information from the scene is lost as stated in \cite{pspnet, deeplabv2, deeplabv3}. 
%As a consequence of the use of relatively small dilation rate values compared to these works \cite{deeplabv2, deeplabv3}, the model might fail to utilize optimal contextual information. 
%
%As it is stated here \cite{pspnet}, the amount of contextual information that we use in DCNN represented by the size of the receptive field. 
%
As a turn around, we employ global average pooling \cite{pspnet} which can retain the contextual information making the feature maps more responsive with context to nearby objects (eg. pedestrian near grass or a pedestrian near a car).
The input of fusion block is fed to global average pooling which generated the average feature maps.
We employ summation to merge the outputs of all pyramid level convolution operations.
The outputs from pyramid pooling and global average pooling are concatenated, and the resulting features is passed to $1 \times 1$ convolution to produce final \textit{logits} (class-wise probabilistic map along the depth channel of the feature map). 
We use bilinear interpolation of \textit{logits} to get full resolution for our prediction maps.
%
%Motivated by recently published paper \cite{deeplabv3}, we add batch normalization in pyramid pooling. 
%
The combination of dilation rate \{2, 4, 8, 16\}, global average pooling along with fusion from the depth branch, we have proven that our results are better when recognizing small objects compared to other methods. This can be verified from the quantitative and qualitative evaluations in the Section \ref{section:experiments}.
To summarize, our propose model consists of four parts. 
First part is feature extractors which is composed by two branches of Atrous ResNet-101 to extract features of RGB and depth images. 
Second part is fusion block that fuses feature maps from both RGB and depth. 
Third part is global average pooling which is useful to retain the contextual information in the feature maps.
The last part consists of pyramid pooling and final prediction. 
The pyramid pooling is composed of five convolution layers that are arranged in a parallel order. 
%
%Then $1 \times 1$ convolution layer is applied to the resulting output to produce final \textit{logits}. 
%
%To obtain a full resolution prediction, we employ bilinear interpolation to upsample final \textit{logits}.
%

\subsection{Implementation Details}
We use pre-trained ResNet-101 \cite{resnet} on ImageNet \cite{imagenet} for feature maps generation in RGB and depth branches separately. 
The original ResNet-101 requires three channels (RGB) image as an input. 
Therefore, in our depth branch, we modify the first convolution layer to accept one channel image.
We initialize the weights for the first convolution layer in depth branch by averaging the weights of the first convolution layer in RGB branch along the channel dimension.
In contrast with the original ResNet-101 \cite{resnet} we remove the first $7 \times 7$ convolution layer and change to three $3 \times 3$ convolution layers as in \cite{rsp, DUC}. 
% Added Dawud
This modification keeps the receptive field similar to additional parameters to learn and makes the network deeper.
We adjust the $\textit{output\_stride} = 8$ by adding dilation $rate = 2$ and $rate = 4$ in the last two residual blocks (res4 and res5 in our naming notation) respectively (see Figure \ref{fig:models}).
We upsample the top-most feature maps with bilinear interpolation by a factor of eight to make the final prediction of full resolution. 
Our implementation is built on top of the open-source code provided by Kong and Fowlkes \cite{rsp} which uses MatConvNet \cite{matconvnet} framework.

\subsection{Training Protocol}
We train our models with the following procedure:
Initially, we train RGB and depth branch separately. 
% NOT NEEDED
%Then, we take Atrous ResNet-101 from depth model without pyramid pooling and final prediction.
%
We insert fusion block before pyramid pooling to combine feature maps generated from both Atrous ResNet-101 from the RGB and depth trained models, which is then passed to pyramid pooling and later for final prediction. 
Finally, we freeze Atrous ResNet-101 for training Fusion Block and Pyramid Pooling to get our final prediction. 
This is achieved by setting the learning rate of the freezing layers to zero.
We employ "poly" learning rate policy, in which a base learning rate is multiplied by $(1-\frac{iter}{maxiter})^{momentum}$.
We set the base learning rate to $5\times10^{-5}$ and $momentum = 0.9$. 
Due to limited GPU memory and large image resolution, we set the batch size to one.
Atrous convolution requires large cropping size to make dilation rate effective \cite{deeplabv3}.
Therefore, we randomly crop the input images to $720 \times 720$ during training on Cityscapes \cite{cityscapes}.
%and  $560\times 560$ for CARLA \cite{carla} dataset.
%
%This size is not too big for our GPU-model and also it is not too small for our dilation rate configuration. 
%
%According to \cite{deeplabv3}, cropping size has to be big enough to make dilation rate effective, or else the filters with large dilation rate are mostly applied to the padded region. 
%
We set the momentum to $0.9$ and weight decay to $0.0005$.
%
%For CARLA dataset \cite{carla}, we set to $560\times560$, since the image size that we generated is only $800\times600$.
%
For data augmentation, we randomly scale the cropped input images with the scale rate between $0.5$ and $2.0$ and also perform left-right flipping. 
In the case of training for the RGB model, we also add color jittering. 
We trained our model for total $200$ epochs. 
From epoch $140$ onwards, we change the base learning rate to $5\times 10^{-4}$ and weight decay of pyramid pooling layers to $0.999$.
\begin{table*}[t!]
\caption{ Quantitative evaluation on Cityscapes dataset. The IoU metric is shown in percentage.}
\label{table:objectIoU}
% \small
\begin{center}
\begin{tabular}{|l|c|c|c|c|c|}
\hline
Object & FCN-8s & RefineNet &  RecurrentParsing & DeepLab-v2(ResNet-101) & Ours \\
\hline\hline
road & 97.40 & 98.20 & 98.50 & 97.86 & 98.45\\
\hline
sidewalk & 78.40 & 83.21 & 85.44 & 81.32 & 85.15\\
\hline
building & 89.21 & 91.28 & 92.51 & 90.35 & 92.24 \\
\hline
wall & 34.93 & 47.78 & 54.41 & 48.77 & 47.10\\
\hline
fence & 44.23 & 50.40 & 60.91 & 47.36 & 59.83\\
\hline
pole & 47.41 & 56.11 & 60.17 & 49.57 & 63.12\\
\hline
traffic light & 60.08 & 66.92 & 72.31 & 57.86 & 71.76 \\
\hline
traffic sign & 65.01 & 71.30 & 76.82 & 67.28 & 76.79\\
\hline
vegetation & 91.41 & 92.28 & 93.10 & 91.85 & 93.22 \\
\hline
terrain & 69.29 & 70.32 & 71.58 & 69.43 & 71.80\\
\hline
sky & 93.86 & 94.75 & 94.83 & 94.19 & 94.62 \\
\hline
person & 77.13 & 80.87 & 85.23 & 79.83 & 84.45\\
\hline
rider & 51.41 & 63.28 & 68.96 & 59.84 & 65.66\\
\hline
car & 92.62 & 94.51 & 95.70 & 93.71 & 95.36 \\
\hline
truck & 35.27 & 64.56 & 70.11 & 56.50 & 58.11\\
\hline
bus & 48.57 & 76.07 & 86.54 & 67.49 & 73.70\\
\hline
train & 46.54 & 64.27 & 75.49 & 57.45 & 61.99\\
\hline
motorcycle & 51.56 & 62.20 & 68.30 & 57.66 & 66.82\\
\hline
bicycle & 66.76 & 69.95 & 75.47 & 68.84 & 74.13 \\
\hline
\hline
Mean IoU & 65.3 & 73.6 & 78.2 & 70.4 & 75.4 \\
\hline
%\end{tabular}
\end{tabular}
\end{center}
\end{table*}
\begin{table*}[h!]
\footnotesize
\caption{Quantitative evaluation on CARLA dataset. The IoU metric is shown in percentage.}
\begin{center}
\begin{tabular}{|l|c|c|c|c|c|}
\hline
Object & FCN-8s & RefineNet &  RecurrentParsing & PSPNet & Ours \\
\hline\hline
Buildings &  57.86 & 77.05 &  72.34 & 71.84 & 79.08 \\
\hline
Fences & 14.04 & 28.53 & 20.47 & 25.01 &  21.78 \\
\hline
Pedestrians & 9.49 & 16.60 &  14.94 & 18.45 & 23.94 \\
\hline
Poles & 16.63 & 37.28 &  20.05 & 27.43 & 38.27 \\
\hline
Roads & 79.62 & 90.69 &  84.59 & 81.90 & 88.96 \\
\hline
Sidewalks & 26.83 & 66.14 & 22.47 & 10.11 & 48.45 \\
\hline
Vegetation & 69.16 & 70.81 &  69.99 & 71.49 & 68.87 \\
\hline
Vehicles & 32.73 & 42.35 &  58.68 & 62.93& 64.86 \\
\hline
Walls & 2.49 & 9.67 &  4.17 & 3.43 & 8.78 \\
\hline
Traffic Signs & 11.11 & 26.90 &  30.21 & 30.69 & 27.65 \\
\hline\hline
Mean IoU & 31.99 & 46.60 &  39.79 & 40.33 & 47.06 \\
\hline
%\end{tabular}
\end{tabular}
\end{center}
\label{table:carla}
\end{table*}

\section{EXPERIMENTAL EVALUATION}
\label{section:experiments}
We evaluate our model on Cityscapes \cite{cityscapes}. 
It is a large-scale outdoor scene dataset recorded across $50$ German cities with different seasons, i.e. spring, summer, and fall.
The dataset contains RGB-D pairs of $2975$ (\textit{train}), $500$ (\textit{validation}) and $1525$ (\textit{test}) image sets with pixel-wise \textit{ fine-annotation} labels.
Additionally, $20000$ extra train RGB-D image pairs with \textit{coarse-annotation} labels is provided. 
We also evaluate on synthetic dataset generated from a publicly available driving simulation framework CARLA \cite{carla}. 
CARLA is an open-source simulator for self-driving car, built on Unreal Engine 4 (UE4) \cite{ue4}.
We generated  $5000$ (\textit{train}) and $500$ (\textit{validation}) RGB-D pairs image sets. 
Furthermore, we perform qualitative evaluation on our self-recorded real data Zed dataset. 
We captured Zed dataset using a front-facing zed stereo camera \cite{zed} mounted on a car.

We provide our quantitative results on \textit{test} data for Cityscapes and \textit{validation} data for CARLA. 
We did adjustments in class mapping for CARLA quantitative measurement as follows: 
First, we ignored two classes, i.e. \textit{road line} and \textit{others}, because they do not exist in Cityscapes.
Then we grouped \textit{car, truck} and \textit{bus} classes from Cityscapes to the \textit{vehicles} class in CARLA. 
Other classes from Cityscapes that do not exist in CARLA are set to \textit{unlabeled} and ignored in IoU measurement.
We compare our results to other well-known approaches such as FCN-8s \cite{fcn}, RefineNet \cite{refineNet}, RecurrentParsing \cite{rsp}, DeepLab-v2(ResNet-101) \cite{deeplabv2} and PSPNet \cite{pspnet}. 
We used intersection-over-union (\textit{IoU})
\footnote{
$
IoU = \frac{TP}{TP+FP+FN},
%\label{eq::iou}
$
where TP is true positive, FP is false positive and FN is false negative pixels. 
}
metric for quantitative evaluation.
We average the IoU result across $10$ classes for CARLA \cite{carla} and $19$ classes for Cityscapes \cite{cityscapes}.
%

%\\explain Figures, Tables etc.
\subsection{Qualitative}
In Figure \ref{fig:city_all}, we compare our qualitative results on Cityscapes \cite{cityscapes} with the ground truth and other methods.
It can be seen that our segmentation is close to ground truth.
In comparison to others, we can see that our segmentation stands out of the FCN-8s \cite{fcn}. 
Compare to RefineNet \cite{refineNet}, we perform better in segmenting narrow \textit{building} and far distant \textit{pole} that can be seen in the $1^{st}$ row.

In Figure \ref{fig:carla_all}, we compare our qualitative results on CARLA \cite{carla} with the ground truth and other methods.
In rows $1^{st}-3^{rd}$, our approach correctly segments \textit{side walk} where other methods fail. 
Furthermore, our model correctly segments other objects in the scenes such as \textit{pedestrians, cars, roads, buildings} and \textit{vegetation}. 
One fail case of our segmentation is in $1^{st}$ row, where it fails to segment \textit{poles} and \textit{traffic sign} nearby the buildings.
It can be because of the same texture and color of the poles with the nearby buildings.
Another visual difference from our result with the ground truth is the \textit{sky} segmentation.
In all prediction images, the \textit{sky} is labelled with the blue color, while in ground truth it is labelled with black.
It happened because the \textit{sky} class does not exist in CARLA \cite{carla}.

In Figure \ref{fig:val_zed}, we perform qualitative evaluation on Zed dataset.
Our model generalized well to real-world data without any fine-tuning or parameters adjustment.
We can see that objects such as \textit{buildings, cars, poles} and \textit{vegetations} are segmented correctly even if they are cluttered.
The failed case is in a wide area of \textit{side walks}.
As we can see in the $5^{th}$ column, the wide \textit{side walks} are segmented as a road.
It possibly due to similar appearance and texture with the \textit{road}.

\subsection{Quantitative}
In Table \ref{table:objectIoU}, we perform quantitative evaluation on Cityscapes \cite{cityscapes}.
We achieve $75.49\%$ accuracy, which is comparable to other state-of-the-art methods.
We gained $5\%$ improvement over the baseline model DeepLab-v2(ResNet-101) \cite{deeplabv2}.
%
%With additional depth branch and custom pyramid pooling network, we gained $5\%$ improvement over the baseline model DeepLab-v2 \cite{deeplabv2}.
%
Additionally, from the Table \ref{table:objectIoU}, we can see that our approach achieves better performance for small objects such as \textit{pole, traffic sign, person, car, terrain} and \textit{vegetation}, while still having comparable results for other objects.
For example in the \textit{pole} class, we gain $13.5 \%$ improvement over the baseline DeepLap-v2(ResNet-101) \cite{deeplabv2} and $2.95\%$ over RecurrentParsing \cite{rsp}.

In Table \ref{table:carla}, we perform quantitative evaluation on CARLA \cite{carla}.
We perform inference on \textit{validation} data.
For a fair comparison, we compare our model with other publicly available Cityscapes-trained models without any fine-tuning or parameters adjustment.
Note that our method achieves higher accuracies in several objects such as \textit{buildings, pedestrians, poles} and \textit{vehicles}.
It may be worth mentioning that all methods perform poorly in segmenting \textit{walls} where accuracies are less than $10\%$.
One reason for this poor segmentation can be that the different shape and texture between the \textit{walls} in CARLA 3D model and real-world object.

\section{CONCLUSIONS}

In this paper, we proposed a network to address multi-scale objects in semantic segmentation.
%
%This scale difference occurs due to the distance from a camera point of view.
%
%We showed that this problem can be handled by incorporating depth information along with multi-scale feature generation.
A novel combination of depth and multi-scale pyramid network specifically address the multi scale objects segmentation.
Our evaluation demonstrated that the proposed network gained $5\%$ improvement over the baseline RGB based method and achieve performance comparable to the state-of-the-art on Cityscapes \cite{cityscapes}.
Furthermore, we showed that our model is robust to other unknown test sets such as, synthetic images generated from CARLA \cite{carla} and on real world images captured using Zed stereo camera \cite{zed}.
Future work includes training the network with multiple GPUs to accomodate the batch size greater than one for faster training and more robust learning.
%
%This is beneficial for faster training time and more robust learning in the network.
%
Additionally, Bayesian learning \cite{mcallister2017concrete} can increase the overall safety of Autonomous Vehicle (AV) by jointly learning the segmentation and uncertainties.
%

%\section*{ACKNOWLEDGMENT}
%
%This work was partially funded by the European project \textit{Eyes of Things} under contract number GA643924. 
%
%Furthermore, we want to thank Oliver Wasenm{\"u}ller and Stephan Krau{\ss} for the fruitful discussions.
%

\printbibliography
\end{document}